\documentclass[11pt,a4paper]{article}
\usepackage[hyperref]{emnlp2020}
\usepackage{times}
\usepackage{latexsym}

\usepackage{microtype}

\aclfinalcopy 

\usepackage[utf8]{inputenc}
\usepackage{url}            
\usepackage{booktabs}       
\usepackage{amsmath}
\usepackage{amsfonts}       
\usepackage{nicefrac}       
\usepackage{natbib}
\usepackage{array}
\usepackage{makecell}
\usepackage{multirow}
\usepackage{amsfonts}
\usepackage{graphicx}
\usepackage{float}
\usepackage{caption} 
\usepackage{mathtools}
\usepackage[autostyle]{csquotes}
\usepackage{caption}
\usepackage{subcaption}
\usepackage{wrapfig}
\usepackage{todonotes}

\usepackage{xcolor}
\colorlet{myGreen}{green!40!gray}
\definecolor{myRed}{RGB}{220, 42, 6}

\DeclareMathOperator*{\argmax}{arg\,max}

\newcommand{\NoGap}{\vspace{-5mm}}  
\newcommand{\CellGap}[1][]{\vspace{0.5em}} 

\makeatletter
\newcommand{\thickhline}{%
    \noalign {\ifnum 0=`}\fi \hrule height 1pt
    \futurelet \reserved@a \@xhline
}
\makeatother

\title{Few-Shot Learning for Opinion Summarization}

\author{Arthur Bražinskas$^1$ \quad Mirella Lapata$^1$ \quad Ivan Titov$^{1,2}$ \bigskip\\
  $^1$ILCC, University of Edinburgh \\
  $^2$ILLC, University of Amsterdam \\
\texttt{abrazinskas@ed.ac.uk, \{mlap,ititov\}@inf.ed.ac.uk}
  }

\date{}

\begin{document}

\setlength{\belowdisplayskip}{5pt}
\setlength{\abovedisplayskip}{5pt}

\maketitle
\begin{abstract}

Opinion summarization is the automatic creation of 
text reflecting subjective information expressed in multiple documents, such as 
user reviews of a product.
The task is practically important and has attracted a lot of attention. 
However, due to the high cost of summary production, 
datasets large enough for training supervised models are lacking. 
Instead, the task has been traditionally approached with
extractive methods 
that learn to select text fragments in an unsupervised or weakly-supervised way. 
Recently,
it has been shown that abstractive
summaries, potentially more fluent and better at reflecting 
conflicting information, can also be produced
in an unsupervised fashion. However, these models, not being exposed to actual 
summaries,  fail to capture their essential properties. 
In this work, we show that even a handful of summaries is sufficient to bootstrap generation of the summary text with all expected properties, such as writing style, informativeness, fluency, and sentiment preservation. We start by training a conditional Transformer language model to
generate a new product review given other available reviews of the product.  
The model is also conditioned on review properties that are directly related to summaries; the properties are derived from reviews with no manual effort. In the second stage, we fine-tune a plug-in module that learns to predict property values on a handful of summaries. This lets us switch the generator to the summarization mode. We show on Amazon and Yelp datasets that our approach substantially outperforms previous extractive and abstractive methods in automatic and human evaluation.   
  
\end{abstract}

\section{Introduction}

Summarization of user opinions expressed in online resources, such as blogs, reviews, social media, or internet forums, has drawn much attention due to its potential for various information access applications, such as creating digests, search, and report generation~\cite{hu2004mining,medhat2014sentiment,angelidis2018summarizing}. 

\begin{table}[t!]
    \centering
 	\footnotesize 
    \begin{tabular}{ >{\centering\arraybackslash} m{1.5cm} m{5cm}} 
 \thickhline
    \textbf{Gold} &  \vspace{0.5em} These shoes run \textcolor{red}{true to size}, \textcolor{myGreen}{do a good job supporting the arch of the foot} and \textcolor{blue}{are well-suited for exercise}. They're good looking, \textcolor{blue}{comfortable}, and the sole feels soft and cushioned. Overall they are a nice, \textcolor{orange}{light-weight pair of shoes} and come in a variety of stylish colors.\vspace{0.5em} \\  \thickhline
    \textbf{Ours} & \vspace{0.5em} These running shoes are great! They \textcolor{red}{fit true to size} and are \textcolor{blue}{very comfortable to run around in}. They are \textcolor{orange}{light weight} and \textcolor{myGreen}{have great support}. They run \textcolor{cyan}{a little on the narrow side}, so make sure to \textcolor{cyan}{order a half size larger than normal}.\vspace{0.5em}  \\ \thickhline
    \textbf{Reviews} & 
    \vspace{0.5em} \textcolor{red}{perfect fit for me} ... \textcolor{myGreen}{supply the support that I need} ...  \textcolor{blue}{are flexible and comfortable}  ... $\vert \vert$ 
    ... \textcolor{blue}{It is very comfortable} ... \textcolor{blue}{I enjoy wearing them running} ... $\vert \vert$ ... \textcolor{blue}{running shoes} ... \textcolor{blue}{felt great right out of the box} ... \textcolor{red}{They run true to size}  ... $\vert \vert$ ... \textcolor{blue}{my feet and feel like a dream} ... \textcolor{orange}{Totally light weight} ... $\vert \vert$ ... \textcolor{cyan}{shoes run small} ... \textcolor{red}{fit more true to size} ... \textcolor{red}{fit is great!} ... \textcolor{myGreen}{supports my arch very well} ... $\vert \vert$ ... \textcolor{orange}{They are lightweight}... \textcolor{cyan}{usually wear a size women's 10 ... ordered a 10.5 and the fit is great!}\vspace{0.5em} \\ \thickhline
    \end{tabular}
    \caption{Example summaries produced by our system and an annotator; colors encode its alignment to the input reviews. The reviews are truncated, and delimited with the symbol `$\vert \vert$'. }
    \label{table:ama_front_example_summ2}
\NoGap
\end{table}

 Although significant progress has been observed in supervised summarization in non-subjective single-document context, such as news articles ~\citep{rush2015neural,
  nallapati2016abstractive, paulus2017deep, see2017get,
  liu2018generating}, modern deep learning methods rely on large
amounts of annotated data that are not readily available in the
opinion-summarization domain and expensive to produce. A key 
obstacle making data annotation
expensive is that annotators 
need to consider  multiple input texts
when writing a summary, which is time-consuming. 
Moreover, annotation 
would have to be undertaken for multiple domains as online reviews are inherently multi-domain ~\cite{blitzer2007biographies} and summarization systems can be domain-sensitive~\cite{isonuma2017extractive}.
This suggests that it is unlikely that human-annotated corpora 
large enough for  training deep models  will
 be available.


Recently, a number of unsupervised abstractive multi-document models were introduced (e.g., \textsc{Copycat}; \citealt{bravzinskas2019unsupervised} and \textsc{MeanSum}; \citealt{chu2019meansum}) that are trained on large collections of unannotated product reviews.\footnote{For simplicity, we use the term `product' to refer to both Amazon products and Yelp businesses.} However, unsurprisingly perhaps, since the models are not exposed to the actual summaries, they are unable to learn their key characteristics. For instance, \textsc{MeanSum} \citep{chu2019meansum} is prone to producing summaries that contain a significant amount of information that is unsupported by reviews; \textsc{Copycat} generates summaries that are better aligned with reviews, yet they are limited in detail. Moreover, both systems, are trained mostly on subjectively written reviews, and as a result, tend to generate summaries in the same writing style.

The main challenge in the absence of large annotated corpora lies in successful utilization of scarce annotated resources. 
Unlike recent approaches to language model adaptation for abstractive single-document summarization \citep{hoang2019efficient, raffel2019exploring} that utilize hundreds of thousands of summaries, our two annotated datasets consist of only 60 and 100 annotated data-points. 
It was also observed that a naive fine-tuning of multi-million parameter models on small corpora leads to rapid over-fitting and poor generalization \citep{vinyals2016matching, finn2017model}. 
In this light, we propose a few-shot learning framework and demonstrate that even a tiny number of annotated instances is sufficient to bootstrap generation of the formal summary text that is both informative and fluent (see Table \ref{table:ama_front_example_summ2}). 
To the best of our knowledge, this work is the first few-shot learning approach 
applied to summarization. 


In our work, we observe that reviews
in a large unannotated collection vary a lot; for example, they differ in style, the level of detail, or how much they diverge from other reviews of the product in terms of content and overall sentiment. We refer to individual review characteristics and their relations to other reviews as \textit{properties} \citep{ficler-goldberg-2017-controlling}. While reviews span a large range of property values, only a subset of them is appropriate
for summaries. For example, summaries should be close to the product's reviews in content, avoid using the first-person pronouns and agree with the reviews in sentiment. Our approach starts with estimating a property-aware model on a large collection of reviews and then adapts the model using a few annotator-created summaries, effectively switching the generator to the summarization regime. As we demonstrate in our experiments, the summaries do not even have to come from the same domain. 

More formally, we estimate a text model
on a dataset of reviews; the generator is a Transformer conditional language model (CLM) that is trained with a `leave-one-out' objective \citep{besag1975statistical, bravzinskas2019unsupervised} by attending to other reviews of the product.
We define properties of unannotated data that are directly related to the end task of summarization. Those properties are easy to derive from reviews, and no extra annotation effort is required.  
The CLM is conditioned on these properties in training. The properties encode partial information about the target review that is being predicted. 
We capitalize on that by fine-tuning parts of the model jointly with a tiny \textit{plug-in network} on a handful of human-written summaries. The plug-in network
is trained to output property values that make the summaries likely under the trained CLM. The plug-in has less than half a percent of the original model's parameters, and thus is less prone to over-fitting on small datasets. Nevertheless, it can successfully learn to control dynamics of a large CLM by providing property values that force generation of summaries. We shall refer to the model produced using the procedure as \textbf{Few} Shot \textbf{Sum}marizer (\textsc{FewSum}).


We evaluate our model against both extractive and abstractive methods on Amazon and Yelp human-created summaries. Summaries generated by our model are substantially better than those produced by competing methods, as measured by automatic and human evaluation metrics on both datasets. Finally, we show that it allows for successful cross-domain adaption. Our contributions can be summarized as follows: 

\begin{itemize}
    \item we introduce the first few-shot learning framework for abstractive opinion summarization;
    \item we demonstrate that the approach substantially outperforms extractive and abstractive models, both when measured with automatic metrics and in human evaluation;
    \item we release datasets with abstractive summaries for Amazon products and Yelp businesses.\footnote{Both the code and datasets are available at: \url{https://github.com/abrazinskas/FewSum}} 
\end{itemize}

\section{Unsupervised Training}
\label{sec:unsup_training}

 \begin{figure*}[t!]
     \centering
     \includegraphics[width=1\textwidth]{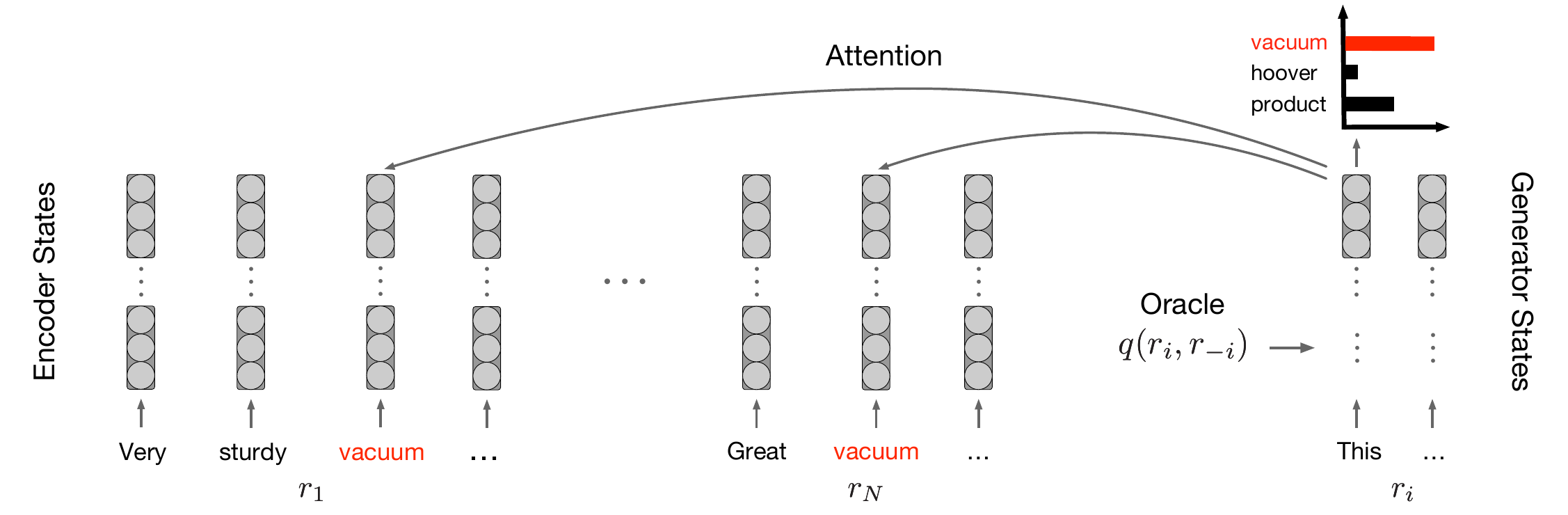}
     \caption{Illustration of the \textsc{FewSum} model that uses the leave-one-out objective. Here predictions of the target review $r_i$ is performed by conditioning on the encoded source reviews $r_{-i}$. The generator attends the last encoder layer's output to extract common information (in \textcolor{myRed}{red}). Additionally, the generator has partial information about $r_i$ passed by the oracle $q(r_i, r_{-i})$.}
     \label{fig:model_general}
 \end{figure*}
 
User reviews about an entity (e.g., a product) are naturally inter-dependent. For example, knowing that most reviews are negative about a product's battery life, it becomes more likely that the next review will also be negative about it. To model inter-dependencies, yet to avoid intractabilities associated with undirected graphical models \citep{koller2009probabilistic}, we use the leave-one-out setting \citep{ besag1975statistical, bravzinskas2019unsupervised}.  

Specifically, we assume access to a large corpus of user text reviews, which are arranged as $M$ groups $\{r_{1:N}\}_{j=1}^M$, where $r_{1:N}$ are reviews about a particular product that are arranged as a \textit{target review} $r_i$ and $N\!-\!1$ \textit{source reviews} $r_{-i}=\{r_1, ..., r_{i-1}, r_{i+1}, ..., r_{N}\}$. Our goal is to estimate the conditional distribution $r_i | r_{-i}$ by optimizing the parameters $\theta$ as shown in Eq. \ref{eq:unsup_objective}. 
\begin{equation}
\begin{aligned}
\theta^* &= \argmax_{\theta} \dfrac{1}{M \; N} \sum_{j=1}^M \sum_{i=1}^N\log p_{\theta}(r_i^j|r_{-i}^j) \\
     &= \argmax_{\theta} \dfrac{1}{M \; N} \sum_{j=1}^M \sum_{i=1}^N\log G_{\theta}(r_i^j| E_{\theta}(r_{-i}^j))
\end{aligned} 
\label{eq:unsup_objective}
\end{equation}


Our model has an encoder-generator Transformer architecture \citep{vaswani2017attention}, where the encoder $E_{\theta}$ produces contextual representations of $r_{-i}$ that are attended by the generator $G_{\theta}$, which in-turn is a conditional language model predicting the target review $r_i$, estimated using teacher-forcing \citep{williams1989learning}.  
An illustration is presented in Fig. \ref{fig:model_general}.

The objective lets the model exploit common information across reviews, such as rare brand names or aspect mentions. For example, in Fig. \ref{fig:model_general}, the generator can directly attend to the word \textit{vacuum} in the source reviews to increase its prediction probability. 

Additionally, we condition on partial information about the target review $r_i$ using an oracle $q(r_i, r_{-i})$ as shown in Eq.  \ref{eq:unsup_objective_final}.
\begin{equation}
\dfrac{1}{M \; N} \sum_{j=1}^M \sum_{i=1}^N\log G_{\theta}(r_i^j| E_{\theta}(r_{-i}^j), q(r_i^j, r_{-i}^j))
\label{eq:unsup_objective_final}
\end{equation}
We refer to this partial information as \textit{properties} \citep{ficler-goldberg-2017-controlling}, which correspond to text characteristics of $r_i$ or relations between $r_i$ and $r_{-i}$. For example, one such property can be the ROUGE score \citep{lin2004rouge} between $r_i$ and $r_{-i}$, which indicates the degree of overlap between $r_i$ and $r_{-i}$. In Fig. \ref{fig:model_general}, a high ROUGE value can signal to the generator to attend the word \textit{vacuum} in the source reviews instead of predicting it based on language statistics. Intuitively, while the model observes a wide distribution of ROUGE scores during training on reviews, during summarization in test time we can achieve a high degree of input-output text overlap by setting the property to a high value. We considered properties that are listed below. 

\textit{Content Coverage}: ROUGE-1, ROUGE-2, and ROUGE-L F1 scores between $r_i$ and $r_{-i}$ signal to $G_{\theta}$ how much to rely on syntactic information in $r_{-i}$ during prediction of $r_i$. \textit{Writing Style}: as a proxy for formal and informal writing styles, we compute pronoun counts, and create a distribution over three points of view. We also added an additional class for cases with no pronouns, see Appendix~\ref{sec:pov_prop} for details and examples; \textit{Rating Deviation}: we compute the difference between the $r_i$'s rating and the average $r_{-i}$ rating; \textit{Length Deviation}: we similarly compute the difference between the $r_i$'s length and the average length of $r_{-i}$, in terms of tokens. 


\subsection{Novelty Reduction}
\label{sec:nov_red}

While summary and review generation are technically similar, there is an important difference that needs to be addressed. Reviews are often very diverse, so when a review is predicted, the generator often needs to predict content that is not present in source reviews.
On the other hand, when a summary is predicted, its semantic content always matches the content of the source reviews. To address this discrepancy, in addition to using the ROUGE scores, as was explained previously, we introduce a \textit{novelty reduction} technique, which is similar to label smoothing \citep{pereyra2017regularizing}.

Specifically, we add a regularization term $\mathcal{L}$, scaled by $\lambda$, that is applied to word distributions produced by the generator $G_{\theta}$ as shown in Eq.~\ref{eq:unsupervised_objective_final_w_penalty}. 
\begin{equation}
\begin{aligned}
&\dfrac{1}{M \; N} \sum_{j=1}^M \sum_{i=1}^N \left[ \log G_{\theta}(r_i^j| E_{\theta}(r_{-i}^j), q(r_i^j, r_{-i}^j))  \right.\\
&- \left. \vphantom{} \lambda \mathcal{L}(G_{\theta}(r_i^j| E_{\theta}(r_{-i}^j), q(r_i^j, r_{-i}^j))\right]
\end{aligned}
\label{eq:unsupervised_objective_final_w_penalty}
\end{equation}
It penalizes assigning the probability mass to words not appearing in $r_{-i}$, as shown in Eq. \ref{eq:nov_pen}, and thus steers towards generation of text that is more grounded in content of $r_{-i}$.
\begin{equation}
\begin{aligned}
    &\mathcal{L}(G_{\theta}(r_i | r_{-i}, q(r_i, r_{-i}))) = \\
    &\sum_{t=1}^{T} \sum_{w \not \in V(r_{-i})} G_{\theta}(W_t=w | r_i^{1:t-1}, r_{-i}, q(r_i, r_{-i}))
    \label{eq:nov_pen}
\end{aligned}
\end{equation}
Here, $T$ is the length of $r_i$, and the inner sum is over all words that do not appear in the word vocabulary of $r_{-i}$. Intuitively, in Fig. \ref{fig:model_general}, the penalty could reduce the probability of the word \textit{hoover} to be predicted as it does not appear in the source reviews.

\section{Summary Adaptation}
\label{sec:summ_adaptation}
Once the unsupervised model is trained on reviews, our task is to adapt it to generation of summaries. Here, we assume access to a small number of annotator-written summaries $(s^k, r^k_{1:N})_{k=1}^K$ where $s$ is a summary for $r_{1:N}$ input reviews.
As we will show in Sec.~\ref{sec:alt_fine_tune}, naive fine-tuning of the unsupervised model on a handful of annotated data-points leads to poor generalization. Instead, we capitalize on the fact that the generator $G_{\theta}$ has observed a wide range of property values associated with $r_i$ during the unsupervised training phase. Intuitively, some combinations of property values drive it into generation of text that has qualities of a summary while others of a review. However, we might not know values in advance that are necessary for generation of summaries. Furthermore, $q(r_i, r_{-i})$ cannot be applied at test time as it requires access to target texts. In the following section, we describe a solution that switches the generator to the summarization mode relying only on input reviews.  


\subsection{Plug-in Network}
\label{sec:plugin_init}
 We start by introducing a parametrized \textit{plug-in network} $p_{\phi}(r_{-i})$ that yields the same types of properties as $q(r_i, r_{-i})$. From a practical perspective, the plug-in should be input-permutation invariant and allow for an arbitrary number of input reviews \citep{zaheer2017deep}. Importantly, the trainable plug-in can have a marginal fraction of the main model's parameters, which makes it less prone to over-fitting when trained on small datasets. We initialize the parameters  of $p_{\phi}(r_{-i})$ by matching its output to $q(r_i, r_{-i})$ on the unannotated reviews. Specifically, we used a weighted combination of distances as shown for one group of reviews in Eq.~\ref{eq:plugin_training}.
 \begin{equation}
\sum_{i=1}^N \sum_{l=1}^L \alpha^l D^l(p_{\phi}(r_{-i})^l, q(r_i, r_{-i})^l)
\label{eq:plugin_training}
\end{equation}

Here, $D^l(p_{\phi}(r_{-i})^l, q(r_i, r_{-i})^l)$ is a distance for the property $l$, and $\alpha^l$ is an associated weight. Specifically, we used L1 norm for \textit{Content Coverage}, \textit{Rating and Length Deviations}, and Kullback-Leibler divergence for \textit{Writing Style}. 

For the plug-in network, we employed a multi-layer feed-forward network with multi-head attention modules over the encoded states of the source reviews at each layer, followed by a linear transformation, predicting property values. Note that the encoder is shared with the main model.

\subsection{Fine-Tuning}
\label{sec:fine_tuning}
Unsurprisingly, perhaps, the network $p_{\phi}$ being initialized on unannotated reviews inherits a strong bias towards outputting property values resulting in generation of reviews, which
should not be appropriate for generating summaries.
Fortunately, due to the simplicity of the chosen properties, it is possible to fine-tune $p_{\phi}$ to match the output of $q$ on the annotated data $(s^k, r^k_{1:N})_{k=1}^K$ using Eq.~\ref{eq:plugin_training}. 

An alternative is to optimize the plug-in to directly increase the likelihood of summaries under $G_{\theta}$ while keeping all other parameters fixed.\footnote{We explored that option, and observed that it works similarly, yet leads to a slightly worse result.} 



As the generator is trained on unannotated reviews, it might not encounter a sufficient amount of text that is written as a summary, and that highly overlaps in content with the input reviews. We address that by unfreezing the attention module of $G_{\theta}$ over input reviews and the plug-in $p_{\phi}$, and by maximizing the likelihood of summaries:
\begin{equation}
    \dfrac{1}{K} \sum_{k=1}^K \left[ \log G_{\theta}(s^k| E_{\theta}(r_{1:N}^k), p_{\phi}(r_{1:N}^k)) \right]
    \label{eq:fine_tuning_obj}
\end{equation}
This allows the system to learn an interaction between $G_{\theta}$ and $p_{\phi}$. For example, what property values are better associated with summaries and how $G_{\theta}$ should better respond to them.



\section{Experimental Setup}
\label{sec:exp_setup}

\subsection{Dataset}
\label{sec:data}
For training we used customer reviews from Amazon \citep{he2016ups} and Yelp.\footnote{\url{https://www.yelp.com/dataset/challenge}}  From the Amazon reviews we selected 4 categories: \textit{Electronics}; \textit{Clothing, Shoes and Jewelry}; \textit{Home and Kitchen}; \textit{Health and Personal Care}. We used a similar pre-processing schema as in \cite{bravzinskas2019unsupervised}, details are presented in Appendix~\ref{sec:dataset_prep}. For training, we partitioned business/product reviews to the groups of 9 reviews by sampling without replacement. Thus, for unsupervised training in Sec.~\ref{sec:unsup_training}, we conditioned on 8 reviews for each target review. The data-statistics are shown in Table~\ref{table:data_stats}. 

\begin{table}
\centering
    \begin{tabular}{ c  c  c  c }     \thickhline
    Dataset & Training & Validation \\ \thickhline
    Yelp & 38,913/1,016,347 & 4,324/113,886\\ 
    Amazon & 182,932/3,889,782 & 9,629/205,992\\ \thickhline
    \end{tabular}
    \caption{Data statistics after pre-processing. The format in the cells is Businesses/Reviews and Products/Reviews for Yelp and Amazon, respectively.}
    \label{table:data_stats}
\NoGap
\end{table}

We obtained 480 human-written summaries (180 for Amazon and 300 for Yelp) for 8 reviews each, using Amazon Mechanical Turk (AMT). Each product/business received 3 summaries, and averaged ROUGE scores are reported in the following sections. Also, we reserved approximately $1\over3$ for testing and the rest for training and validation. The details are in Appendix~\ref{sec:eval_data_split}.




\subsection{Experimental Details}


For the main model, we used the Transformer architecture \citep{vaswani2017attention} with trainable length embeddings and shared parameters between the encoder and generator \citep{raffel2019exploring}. Subwords were obtained with BPE \citep{sennrich2015neural} using 32000 merges. Subword embeddings were shared across the model as a form of regularization \citep{press2016using}. For a fair comparison, we approximately matched the number of parameters to \textsc{Copycat} \citep{bravzinskas2019unsupervised}. We randomly initialized all parameters with Glorot \citep{glorot2010understanding}. For the plug-in network, we employed a multi-layer feed-forward network with multi-head attention modules over encoded states of the source review. After the last layer, we performed a linear projection to compute property values. Further, parameter optimization was performed using Adam \citep{kingma2014adam}, and beam search with n-gram blocking \citep{paulus2017deep} was applied to our model and Copycat for summary generation. 

All experiments were conducted on 4 x GeForce RTX 2080 Ti.  

\subsection{Hyperparameters}
\label{sec:hyper}
Our parameter-shared encoder-generator model used a 8-head and 6-layer Transformer stack. Dropout in sub-layers and subword embeddings dropout was both set to 0.1, and we used 1000 dimensional position-wise feed-forward neural networks. We set subword and length embeddings to 390 and 10 respectively, and both were concatenated to be used as input. For the plug-in network, we set the output dimension to 30 and internal feed-forward network hidden dimensions to 20. We used a stack of 3 layers, and the attention modules with 3 heads at each layer. We applied 0.4 internal dropout and 0.15 attention dropout. Property values produced by the plug-in or oracle were concatenated with subword and length embeddings and linearly projected before being passed to the generator. In total, our model had approximately 25M parameters, while the plug-in network only 100K (i.e., less than 0.5 \% of the main model's parameters).

In all experiments, the hyperparameter tuning was performed based on the ROUGE-L score on Yelp and Amazon validation sets.

\subsection{Baselines}
\label{sec:baselines}

\textsc{LexRank} \citep{erkan2004lexrank} is an unsupervised extractive graph-based algorithm selecting sentences based on graph centrality. Sentences represent nodes in a graph whose edges have weights denoting similarity computed with tf-idf. 

\textsc{MeanSum} is an unsupervised abstractive summarization model~\cite{chu2019meansum} that treats a summary as a discrete latent state of an autoencoder. 
The model is trained in a multi-task fashion with two objectives, one for prediction of reviews and the other one for summary-reviews alignment in the semantic space using the cosine similarity.

\textsc{Copycat} is the state-of-the-art unsupervised abstractive summarizer~\cite{bravzinskas2019unsupervised} that uses continuous latent representations to model review groups and individual review semantics. It has an implicit mechanism for novelty reduction and uses a copy mechanism.

As is common in the summarization literature, we also employed a number of simple summarization baselines. First, the \textsc{clustroid} review was computed for each group of reviews as follows. We took each review from a group and computed ROUGE-L with respect to all other reviews. The review with the highest ROUGE score was selected as the clustroid review. Second, we sampled a \textsc{random} review from each group to be used as the summary. Third, we constructed the summary by selecting the \textit{leading sentences} (\textsc{Lead}) from each review of a group.

\section{Evaluation Results}
\label{sec:eval}

\paragraph{Automatic Evaluation}

\begin{table}[h!]
\centering
    \begin{tabular}{  l  c  c  c }     \thickhline
     & R1 & R2 & RL \\ \thickhline
    FewSum & \textbf{0.3356} & \textbf{0.0716} & \textbf{0.2149}\\
    Copycat & 0.2785 & 0.0477 & 0.1886 \\  
    MeanSum & 0.2663 & 0.0489 & 0.1711\\ 
    LexRank & 0.2772 & 0.0506 & 0.1704 \\ \thickhline
    Clustroid & 0.2716 & 0.0361 & 0.1677 \\ 
    Lead & 0.2700 & 0.0492 & 0.1495 \\ 
    Random & 0.2500 & 0.0382 & 0.1572 \\ \thickhline
    \end{tabular}
    \caption{ROUGE scores on the Amazon test set.}
    \label{table:amazon_rouge}
\end{table}

\begin{table}[h!]
\centering
    \begin{tabular}{  l  c  c  c }     \thickhline
     & R1 & R2 & RL \\ \thickhline
    FewSum & \textbf{0.3729} & \textbf{0.0992} & \textbf{0.2276}\\
    Copycat & 0.2812 & 0.0589  & 0.1832  \\  
    MeanSum & 0.2750 & 0.0354 & 0.1609 \\ 
    LexRank & 0.2696 &  0.0493 & 0.1613 \\ 
    \thickhline
    Clustroid & 0.2890 & 0.0490 & 0.1800 \\ 
    Lead & 0.2620 & 0.0457 & 0.1432 \\ 
    Random & 0.2148 & 0.0259 & 0.1387 \\ \thickhline
    \end{tabular}
    \caption{ROUGE scores on the Yelp test set.}
    \label{table:yelp_rouge}
\NoGap
\end{table}

We report ROUGE F1 score \citep{lin2004rouge} based evaluation results on the Amazon and Yelp test sets in Tables~\ref{table:amazon_rouge} and~\ref{table:yelp_rouge}, respectively. The results indicate that our model outperforms abstractive and extractive methods on both datasets. Also, the results are supported by qualitative improvements over other models, see examples in the Appendix.

\begin{table*}[t]
\centering
\begin{tabular}{ l  c  c  c  c  c } \thickhline
       & Fluency & Coherence & Non-Redundancy & Informativeness & Sentiment \\ \thickhline
       FewSum & \textbf{0.1000} & \textbf{0.1429} & \textbf{0.1250} & \textbf{0.2000} & \textbf{0.3061} \\ 
       Copycat & -0.1765 & -0.5333 & -0.2727 & -0.7455 & -0.7143 \\ 
       LexRank &  -0.4848 & -0.5161 & -0.5862 & -0.3488 & -0.0909 \\ 
       \thickhline
       Gold & 0.5667 &  0.6364 & 0.6066 & 0.6944 & 0.4138 \\ \thickhline
\end{tabular}
    \caption{Human evaluation results in terms of the Best-Worst scaling on the Amazon test set.}
    \label{table:ama_human_eval}
\end{table*}

\paragraph{Best-Worst Scaling}

\begin{table*}[t]
\centering
\begin{tabular}{ l  c  c  c  c  c } \thickhline
       & Fluency & Coherence & Non-Redundancy & Informativeness & Sentiment \\ \thickhline
       FewSum & \textbf{0.1636} & \textbf{0.1429} & 0.0000 &  \textbf{0.3793} & \textbf{0.3725} \\ 
       Copycat & -0.2000 & -0.0769 & \textbf{0.1053} & -0.4386 & -0.2857 \\ 
       LexRank & -0.5588 & -0.5312 & -0.6393 & -0.6552 & -0.4769 \\ 
       \thickhline
       Gold & 0.5278 & 0.3784 & 0.4795 & 0.6119 & 0.4118 \\ \thickhline
\end{tabular}
    \caption{Human evaluation results in terms of the Best-Worst scaling on the Yelp test set.}
    \label{table:yelp_human_eval}
\end{table*}

We performed human evaluation with the Best-Worst scaling \citep{louviere1991best, louviere2015best, kiritchenko2017capturing} on the Amazon and Yelp test sets using the AMT platform. We assigned multiple workers to each tuple containing summaries from \textsc{Copycat}, our model, \textsc{LexRank}, and human annotators. The judgment criteria were the following: \textit{Fluency}, \textit{Coherence}, \textit{Non-redundancy}, \textit{Informativeness}, \textit{Sentiment}. Details are provided in Appendix \ref{app:best_worst}.

For every criterion, a system’s score is computed as the percentage of times it was selected as best, minus the percentage of times it was selected as worst \citep{orme2009maxdiff}. The scores range from -1 (unanimously worst) to +1 (unanimously best).

The results are presented in Tables~\ref{table:ama_human_eval} and~\ref{table:yelp_human_eval} for Amazon and Yelp, respectively. On the Amazon data, they indicate that our model is preferred across the board over the baselines. \textsc{Copycat} is 
preferred over \textsc{LexRank} in terms of fluency and non-redundancy, yet it shows worse results in terms of informativeness and overall sentiment preservation. In the same vein, on Yelp in Table~\ref{table:yelp_human_eval} our model outperforms the other models. 


All pairwise differences between our model and other models are statistically significant at $p < 0.05$, using post-hoc HD Tukey tests. The only exception is non-redundency on Yelp when comparing our model and \textsc{Copycat} (where our model shows a slightly lower score). 


\paragraph{Content Support}
As was observed by \citet{falke2019ranking, tay2019red, bravzinskas2019unsupervised}, the ROUGE metric can be insensitive to hallucinating facts and entities. We also investigated how well generated text is supported by input reviews. We split summaries generated by our model and \textsc{Copycat} into sentences. Then for each summary sentence, we hired 3 AMT workers to judge how well content of the sentence is supported by the reviews. Three following options were available. \textit{Full
  support}: all the content is reflected in the reviews;
\textit{Partial support}: only some content is reflected in the
reviews; \textit{No support}: content is not reflected in the reviews.

The results are presented in Table~\ref{table:content_support}. Despite not using the copy mechanism, that is beneficial for fact preservation \citep{falke2019ranking} and generation of more diverse and detailed summaries (see Appendix), we score on par with \textsc{Copycat}.
\begin{table}[t]
\centering
\begin{tabular}
{ l  c  c  c} \thickhline
    & Full (\%) & Partial (\%)  & No (\%) \\ \thickhline
FewSum & 43.09 & \textbf{34.14} & \textbf{22.76} \\
Copycat & \textbf{46.15} & 27.18 & 26.67 \\ \thickhline
\end{tabular}
\caption{Content support on  the Amazon test set.}
\label{table:content_support}
\NoGap
\end{table}


\section{Analysis}
\label{sec:analysis}

\subsection{Alternative Adaptation Strategies}
\label{sec:alt_fine_tune}
We further explored alternative utilization approaches of annotated data-points, based on the same split of the Amazon summaries as explained in Sec. \ref{sec:data}. First, we trained a model in an unsupervised learning setting (\textsc{USL}) on the Amazon reviews with the leave-one-out objective in Eq. \ref{eq:unsup_objective}. In this setting, the model has neither exposure to summaries nor the properties, as the oracle $q(r_i, r_{-i})$ is not used. Further, we considered two alternative settings how the pre-trained unsupervised model can be adapted on the gold summaries. In the first setting, the model is fine-tuned by predicting summaries conditioned on input reviews (\textsc{USL+F}). In the second one, similar to \citet{hoang2019efficient}, we performed adaptation in a multi-tasking learning (\textsc{MTL}) fashion. Here, \textsc{USL} is further trained on a mixture of unannotated corpus review and gold summary batches with a trainable embedding indicating the task.\footnote{We observed that the 1:1 review-summary proportion works the best.} The results are presented in Table~\ref{table:alt_amazon_rouge}.

First, we observed that \textsc{USL} generates summaries that get the worst ROUGE scores. Additionally, the generated text tends to be informal and substantially shorter than an average summary, we shall discuss that in Sec.~\ref{sec:summ_char}.
Second, when the model is fine-tuned on the gold summaries (\textsc{USL+F}), it noticeably improves the results, yet they are substantially worse than of our proposed few-shot approach. It can be explained by strong influence of the unannotated data stored in millions of parameters that requires more annotated data-points to overrule. Finally, we observed that \textsc{MTL} fails to decouple the tasks, indicated by only a slight improvement over \textsc{USL}.

\begin{table}[t!]
\centering
    \begin{tabular}{  l  c  c  c }     \thickhline
     & R1 & R2 & RL \\ \thickhline
    FewSum & 0.3356 & 0.0716 & 0.2149 \\ \thickhline
    MTL & 0.2403 & 0.0435 & 0.1627 \\ 
    USL+F & 0.2823 & 0.0624 & 0.1964 \\
    USL & 0.2145 & 0.0315 & 0.1523 \\ \thickhline
    Random & 0.2500 & 0.0382 & 0.1572 \\ \thickhline
    \end{tabular}
    \caption{ROUGE scores on the Amazon test set for alternative summary adaptation strategies.}
    \label{table:alt_amazon_rouge}
\end{table}

\subsection{Influence of Unannotated Data}
\label{sec:summ_char}
We further analyzed how  plain fine-tuning on summaries differs from our approach in terms of capturing summary characteristics. For comparison, we used \textsc{USL} and \textsc{USL+F}, which are presented in Sec.~\ref{sec:alt_fine_tune}. Additionally, we analyzed unannotated reviews from the Amazon training set. Specifically, we focused on text formality and the average word count difference (\textit{Len}) from the gold summaries in the Amazon test set. As a proxy for the former, we computed the marginal distribution over points of view (POV), based on pronoun counts; an additional class (\textit{NoPr}) was allocated to cases of no pronouns. The results are presented in Table~\ref{table:text_char}.


\begin{table}[t]
\centering
\begin{tabular}
{ l |c c c c |c } \thickhline
    & 1st & 2nd & 3rd & NoPr & Len\\ \thickhline
Gold & 0.0 & 1.7 & 60.0 & 38.3 & 0.0\\ \thickhline
FewSum & 0.5 & 1.3 & 83.2 & 15.0 & 3.4 \\ 
USL+F & 29.7 & 0.0 & 45.3 & 25.0 & -28.6\\ 
USL & 56.7 & 0.0 & 43.3 & 0.0 & -32.7\\ \thickhline
    Reviews & 49.0 & 7.3 & 35.6 & 8.1 & -17.6 \\ \thickhline
\end{tabular}
\caption{Text characteristics of generated summaries by different models on the Amazon test set. }
\label{table:text_char}
\NoGap
\end{table}

First, we observed that the training reviews are largely informal (49.0\% and 7.3\% for 1st and 2nd POV, respectively). Unsurprisingly, the model trained only on the reviews (\textsc{USL}) transfers a similar trait to the summaries that it generates.\footnote{As beam search, attempting to find the most likely candidate sequence, was utilized, opposed to a random sequence sampling, we observed that generated sequences had no cases of the 2nd POV pronouns and complete absence of pronouns (NoPr).} On the contrary, the gold summaries are largely formal - indicated by a complete absence of the 1st and a marginal amount of 2nd POV pronouns. Also, an average review is substantially shorter than an average gold summary, and consequently, the generated summaries by \textsc{USL} are also shorter. Example summaries are presented in Table~\ref{table:ama_analysis_summ}.

Further, we investigated how well \textsc{USL+F}, adapts to the summary characteristics by being actually fine-tuned on them. Indeed, we observed that \textsc{USL+F} starts to shift in the direction of the summaries by reducing the pronouns of the 1st POV and increasing the average summary length. Nevertheless, the gap is still wide, which would probably require more data to be bridged. Finally, we observed that our approach adapts much better to the desired characteristics by producing well-formed summary text that is also very close in length to the gold summaries. 

\begin{table}[t!]
    \centering
 	\footnotesize 
    \begin{tabular}{ >{\centering\arraybackslash} m{1.5cm} m{5cm}} 
 \thickhline
  \textbf{Gold} & \vspace{0.5em}These shoes run true to size, do a good job supporting the arch of the foot and are well-suited for exercise. They're good looking, comfortable, and the sole feels soft and cushioned. Overall they are a nice, light-weight pair of shoes and come in a variety of stylish colors. \vspace{0.5em} \\ \thickhline
  \textbf{FewSum} & \vspace{0.5em} These running shoes are great! They fit true to size and are very comfortable to run around in. They are light weight and have great support. They run a little on the narrow side, so make sure to order a half size larger than normal.\vspace{0.5em} \\ \hline
  \textbf{USL+F} & \vspace{0.5em} This is my second pair of Reebok running shoes and they are the best running shoes I have ever owned. They are lightweight, comfortable, and provide great support for my feet. \vspace{0.5em} \\ \hline 
  \textbf{USL} &  \vspace{0.5em} This is my second pair of Reebok running shoes and I love them. They are the most comfortable shoes I have ever worn. \vspace{0.5em} \\ \hline
    \end{tabular}
    \caption{Example summaries produced by models with different adaptation approaches.}
    \label{table:ama_analysis_summ}
\end{table}

\subsection{Cross-Domain}
\begin{table}[t!]
\centering
    \begin{tabular}{  l  c  c }     \thickhline
     Domain & In-domain & Cross-domain \\ \thickhline
     Cloth & 0.2188 & 0.2220 \\
     Electronics & 0.2146 & 0.2136 \\
     Health & 0.2121 & 0.1909 \\
     Home & 0.2139 & 0.2250 \\ \thickhline
     Avg & 0.2149 & 0.2129 \\ \thickhline
    \end{tabular}
    \caption{In and cross domain experiments on the Amazon dataset, ROUGE-L scores are reported.}
    \label{table:ama_cross_domain}
    \NoGap
\end{table}
We hypothesized that on a small dataset, the model primarily learns course-grained features, such as common writing phrases, and their correlations between input reviews and summaries. Also, that they, in principle, could be learned from remotely related domains. We investigated that by fine-tuning the model on summaries that are not in the target domain of the Amazon dataset. Specifically, we matched data-point count for 3/4 domains of training and validation sets to the in-domain Amazon data experiment presented in Sec~\ref{sec:eval}; the test set remained the same for each domain as in the in-domain experiment. Then, we fine-tuned the same model 5 times with different seeds per target domain. For comparison, we used the in-domain model which was used in Amazon experiments in Sec.~\ref{sec:eval}. We computed the average ROUGE-L score per target domain, where overall $\sigma$ was 0.0137. The results are reported in Table~\ref{table:ama_cross_domain}.

The results indicate that the models perform on-par on most of the domains, supporting the hypothesis. On the other hand, the in-domain model shows substantially better results on the \textit{health} domain, which is expected, as, intuitively, this domain is the most different from the rest. 



\section{Related Work}
Extractive weakly-supervised opinion summarization has been an active
area of research. 
\textsc{LexRank}~\citep{erkan2004lexrank} is an unsupervised extractive model.
\textsc{Opinosis}~\citep{ganesan2010opinosis} does not use any supervision and relies on POS tags and redundancies to generate short opinions. However, this approach is not well suited for the generation of
coherent long summaries and, although it can recombine fragments of
input text, it cannot generate novel words and phrases. 
Other earlier approaches \citep{gerani2014abstractive, di2014hybrid} relied on text planners and templates, which restrict the output text. A more recent extractive method of \citet{angelidis2018summarizing} frames the problem as a pipeline of steps with different models for each step. \citet{isonuma2019unsupervised} introduce an unsupervised approach for single product review summarization, where they rely on latent discourse trees. 
The most related unsupervised approach to this work is our own work, \textsc{Copycat}~\citep{bravzinskas2019unsupervised}. Unlike that work, we rely on a powerful generator to learn conditional spaces of text without hierarchical latent variables. Finally, in contract to \textsc{MeanSum}~\citep{chu2019meansum}, our model relies on inductive biases without explicitly modeling of summaries.
A concurrent model \textsc{DenoiseSum}~\citep{amplayo2020unsupervised} uses a syntactically generated dataset of source reviews to train a generator to denoise and distill common information. Another parallel work, \textsc{OpinionDigest}~\citep{suhara2020opiniondigest}, considers controllable opinion aggregation and is a pipeline framework for 
abstractive summary generation. 
Our conditioning on text properties approach is similar to \citet{ficler-goldberg-2017-controlling}, yet we rely on automatically derived properties that associate a target to source, and learn a separate module to generate their combinations. Moreover, their method has not been studied in the context of summarization.


\section{Conclusions}

In this work, we introduce the first to our knowledge few-shot framework for abstractive opinion summarization. We show that it can efficiently utilize even a handful of annotated reviews-summary pairs to train models that generate fluent, informative, and overall sentiment reflecting summaries. We propose to exploit summary related properties in unannotated reviews that are used for unsupervised training of a generator. Then we train a tiny plug-in network that learns to switch the generator to the summarization regime. We demonstrate that our approach substantially outperforms competitive ones, both abstractive and extractive, in human and automatic evaluation. Finally, we show that it also allows for successful cross-domain adaptation.  

\section*{Acknowledgments}

We would like to thank members of Edinburgh NLP group for discussion, and the anonymous reviewers for their valuable feedback.
We gratefully acknowledge the support of the European Research
Council (Titov: ERC StG BroadSem 678254; Lapata: ERC CoG TransModal 681760) and the Dutch National Science Foundation (NWO VIDI 639.022.518).

\newpage
\bibliography{bibl}
\bibliographystyle{acl_natbib}

\newpage
\section{Appendices}
\label{sec:appendix}
 \subsection{Dataset Pre-Processing}
 \label{sec:dataset_prep}
 We selected only Amazon products and Yelp businesses with minimum of 10 reviews, and the minimum and maximum lengths of 20 and 70 words, respectively. Also, popular products/businesses above the $90^{th}$ percentile were removed. From each business/product we sampled 9 reviews without replacement to form groups of reviews. 

\subsection{Evaluation Data Split}
\label{sec:eval_data_split}
From the Amazon annotated dataset, We used 28, 12, 20 products for training, validation, and testing, respectively. On Yelp, we used 30, 30, 40 for training, validation, and testing, respectively. Both the automatic and human evaluation experiments were performed on the test sets.

\subsection{Training Procedure}
\label{sec:training_proc}

First, to speed-up the training phase, we trained an unconditional language model for 13 epoch on the Amazon reviews with the learning rate (LR) set to $5*10^{-4}$. On Yelp we trained it for 27 epochs with LR set to $7*10^{-4}$. The language model was used to initialize both the encoder and generator of the main model.  

Subsequently, we trained the model using Eq. \ref{eq:unsup_objective_final} for 9 epochs on the Amazon reviews  with $6 * 10^{-5}$ LR, and for 57 epochs with LR set to $5*10^{-5}$.
Additionally, we reduced novelty using Eq. \ref{eq:nov_pen} by training the model further for 1 epoch with $10^{-5}$ LR and $\lambda=2$ on Amazon; On Yelp we trained for 4 epochs, with $3*10^{-5}$ LR, and $\lambda=2.5$.

For the plugin network's initialization, as explained in Sec.~\ref{sec:plugin_init}, we performed optimization by output matching with the oracle for 11 epochs on the unannotated Amazon reviews with $1*10^{-5}$ LR. On Yelp we trained for 87 epochs with $1*10^{-5}$
Lastly, we fine-tuned the plugin network on the human-written summaries by output matching with the oracle\footnote{We set rating deviation to 0 as summaries do not have associated human-annotated ratings.}. On the Amazon data for 98 epochs with $7 * 10^{-4}$, and for 62 epochs with $7*10^{-5}$ on Yelp. We set weights to 0.1, 1., 0.08, 0.5 for length deviation, rating deviation, POV, and ROUGE scores, respectively. Then fine-tuned the attention part of the model and the plug-in network jointly for 33 epochs with $1*10^{-4}$ on the Amazon data. And 23 epochs with $1*10^{-4}$ LR on Yelp.

\subsection{Summary Annotation}
\label{sec:summ_annotation}

For summary annotation, we reused 60 Amazon products from \citet{bravzinskas2019unsupervised} and sampled 100 businesses from Yelp. 
We assigned 3 Mechanical Turk workers to each product/business, and instructed them to read the reviews and produce a summary text. We used the following instructions:

\begin{itemize}
    \item The summary should reflect user common opinions expressed in the reviews. Try to preserve the common sentiment of the opinions and their details (e.g. what exactly the users like or dislike). For example, if most reviews are negative about the sound quality, then also write negatively about it.
    \item Please make the summary coherent and fluent in terms of sentence and information structure. Iterate over the written summary multiple times to improve it, and re-read the reviews whenever necessary. 
    \item The summary should not look like a review, please write formally.
    \item Keep the length of the summary reasonably close to the average length of the reviews.
    \item Please try to write the summary using your own words instead of copying text directly from the reviews. Using the exact words from the reviews is allowed but do not copy more than 5 consecutive words from a review. 
\end{itemize}

\subsection{Human Evaluation Setup}
\label{sec:he_setup}

To perform the human evaluation experiments described in Sec~\ref{sec:eval}, we hired workers with 98\% approval rate, 1000+ HITS, Location: USA, UK, Canada, and the maximum score on a qualification test that we had designed. The test was asking if the workers were native English speakers, and was verifying that they correctly understood the instructions of both the best-worst scaling and content support tasks.

\subsection{Best-Worst Scaling Details}
\label{app:best_worst}

We performed human evaluation based on the Amazon and Yelp test sets using the AMT platform. We assigned workers to each tuple containing summaries from \textsc{Copycat}, our model, \textsc{LexRank}, and human annotators. Due to dataset size differences, we assigned 5 and 3 workers to each tuple in the Amazon and Yelp test sets, respectively. We presented the associated reviews in a random order and asked the workers to judge summaries using the Best-Worst scaling (BWS) \citep{louviere1991best, louviere2015best} that is known to produce more reliable results than ranking scales \citep{kiritchenko2017capturing}. The judgment criteria are presented below, where \textit{non-redundancy} and \textit{coherence} were taken from \citet{dang2005overview}.
    \textit{Fluency}: the summary sentences should be grammatically correct, easy to read and understand; 
    \textit{Coherence}: the summary should be well structured and well organized;
     \textit{Non-redundancy}: there should be no unnecessary repetition in the summary;
     \textit{Informativeness}: how much useful information about the product does the summary provide?; 
    \textit{Sentiment}: how well the sentiment of the summary agrees with the overall sentiment of the original reviews?

\subsection{Points of View}
\label{sec:pov_prop}

\begin{table}
    \centering
    \footnotesize
    \begin{tabular}{ >{\centering\arraybackslash} m{1cm} m{5.5cm}} 
     \thickhline
         1st &  \CellGap \vspace{0.2cm}
         I bought this as a gift for my husband.
         \newline \noindent\rule{5.5cm}{0.2pt}\vspace{0.2cm}
         I've been using Drakkar Noir Balm for over twenty years. 
         \newline \noindent\rule{5.5cm}{0.2pt}\vspace{0.2cm}
         I purchased these for my son as a kind of a joke. \CellGap \vspace{0.1cm}\\\thickhline
         2nd & \CellGap \vspace{0.2cm}
         This is the best product you can buy! 
        \newline \noindent\rule{5.5cm}{0.2pt}\vspace{0.2cm}
        You get what you pay for. 
         \newline \noindent\rule{5.5cm}{0.2pt}\vspace{0.2cm}
        Please do yourself a favor and avoid this product.
               \CellGap \vspace{0.1cm}\\ \thickhline
               
         3rd & \CellGap \vspace{0.2cm}
         This is his every work day scent.          \newline \noindent\rule{5.5cm}{0.2pt}\vspace{0.2cm}
               It's very hard to buy the balm separately.          \newline \noindent\rule{5.5cm}{0.2pt}\vspace{0.2cm}

               It smells like Drakkar, but it is hard to find. \CellGap \vspace{0.1cm}
         \\  \thickhline
         No \newline
         Pronouns & \CellGap \vspace{0.2cm} Very nice, not too overpowering.          \newline \noindent\rule{5.5cm}{0.2pt}\vspace{0.2cm}

                 This product has no smell what ever.          \newline \noindent\rule{5.5cm}{0.2pt}\vspace{0.2cm}

                 Nice to use for hardwood floors. \CellGap \vspace{0.1cm} \\  \thickhline
    \end{tabular}
    \caption{Examples of review sentences that contain only pronouns belonging to a specific class.}
    \label{table:POV_examples}
    \NoGap
\end{table}

Summaries differ from reviews in terms of the writing style. Specifically, reviews are predominantly written informally, populated by pronouns such as \textit{I} and  \textit{you}. In contrast, summaries are desirable to be written formally. In this work, we observed that a surprisingly simple way to achieve that is to condition the generator on the distribution over pronoun classes of the target review. We computed pronoun counts and produced the 4 class distributions: 1st, 2nd, 3rd person POV, and `other' in case if no pronouns are present.

Consider the example sentences in Table~\ref{table:POV_examples}. Here one can observe that the sentences of different pronoun classes differ in the style of writing and often the intention of the message: 1st POV sentences tend to provide clues about the personal experience of the user; 2nd POV sentences, on the other hand, commonly convey recommendations to a reader; 3rd POV and `other` sentences often describe aspects and their associated opinions.


\begin{table*}
    \centering
 	\footnotesize 
    \begin{tabular}{  >{\centering\arraybackslash} m{1.5cm} m{12cm}}
 \thickhline
    Gold & \vspace{0.5em} Bennett Medical has poor customer service. Phone calls can take a long time to get answered and leaving voice mails tend to be fruitless. The products are overpriced and take a long time to be refilled. Using this medical supply company can be a hassle.\vspace{0.5em} \\ \hline
    Ours & \vspace{0.5em} This medical supply is the worst medical supply company in the valley. The customer service is horrible, the staff is rude, the wait times are long, and the service reps are not helpful at all. Do not recommend this company to anyone.
          \vspace{0.5em} \\ \hline
    Copycat & \vspace{0.5em} If I could give them 0 stars I would. The customer service is terrible, the staff is extremely rude and helpful. If you're looking for a new provider, this is the place to be. I will never use them again.\vspace{0.5em} \\ \hline
    MeanSum & \vspace{0.5em} Service is horrible, especially the manager. I have a lot of kids but not this place. Two months later I was able to go in and get a new one to go in the next day. They would tell me that they would have to do a lot of our water to be there to get a new one. But this is the first time I have dealt with him and we will never go back again. Thanks for your hard work, and I will never go anywhere else. \vspace{0.5em} \\ \hline
    Lexrank & \vspace{0.5em} Bennett Medical for Cpap supplies are horrible. Never enough staff to answer phone, so you'll need to leave messages. DON'T use this medical supply. If I could give Bennett Medical zero stars I would! Will be moving to another medical supply as soon as I can. \vspace{0.5em} \\ \hline
     \thickhline
    Review 1 & \vspace{0.5em} Bennett Medical for Cpap supplies are horrible. We have waited for three weeks to refill supplies and we are still waiting. This company does not have good customer service, you can only leave messages, and they never call back. If I could give Bennett Medical zero stars I would!\vspace{0.5em} \\ \hline
    Review 2 & \vspace{0.5em} Teachers Health Trust, please look into the practice of the billing and filling of durable services. The mask cushions go for 45 to 50 days because of the lack of communication. The people in charge of billing are very argumentative and lack customer service. I will drop them after annual, because of my insurance obligations.\vspace{0.5em} \\ \hline
    Review 3 & \vspace{0.5em} Fantastic service from Jocelyn at the front desk, we had a really hard time getting the right paperwork together from Drs but she stuck with us and helped us every step of the way, even calling to keep us updated and to update info we might have for her. Thanks Jocelyn.\vspace{0.5em} \\ \hline
    Review 4 & \vspace{0.5em} I hardly ever write reviews, but I'd like to spare someone else from what I experienced. So a warning to the wise... If you like rude incompetent employees, almost an hour long wait for just picking up a phone order, and basically being treated like a second class citizen then look no further than Bennett Medical.\vspace{0.5em} \\ \hline
    Review 5 & \vspace{0.5em} DON'T use this medical supply. Never enough staff to answer phone, so you'll need to leave messages. No return phone calls. I am unable to get my CPAP supplies every quarter without hours of calling / waiting / calling. Poor customer service. Will be moving to another medical supply as soon as I can.\vspace{0.5em} \\ \hline
    Review 6 & \vspace{0.5em} Terrible experience. They have ridiculous price also bad customer services. You can get nebulizer machine around \$50 at amazon, bennet medical charge you almost twice more expensive price. And breathing kit price was unbelievable too. Because of deduction, I had to pay them all out of my pocket whatever they charged. I don't recommand this medical company to anyone.\vspace{0.5em} \\ \hline
    Review 7 & \vspace{0.5em} Good luck getting a phone call back or someone to answer the phone without hanging up immediately. I have called over 20 times left 5 voicemails over the last 30 days, just to refill a mask perscription. This is an ongoing issue that is beyond frustrating. Not trying to destroy this businesses name just want the owners to implement some basic customer service skills.\vspace{0.5em} \\ \hline
    Review 8 & \vspace{0.5em} Always receive friendly customer service whenever we call or go into the location. My questions are always answered and I am very happy with the supplies we get from them. Great people providing a great service! Thank you for all you do!\vspace{0.5em} \\ \thickhline
    \end{tabular}
    \caption{Example summaries produced by different systems on Yelp data.}
\end{table*}


\begin{table*}
    \centering
 	\footnotesize 
    \begin{tabular}{  >{\centering\arraybackslash} m{1.5cm} m{12cm}}
 \thickhline
    Gold & \vspace{0.5em}  It is very clean and nice inside. Everyone is so kind and friendly. They do an amazing job on both nails and pedis. They get it done with speed and precision with a price that is very much affordable. They have the best customer service. \vspace{0.5em} \\ \hline
    Ours & \vspace{0.5em} This nail salon is very clean and the staff is very friendly. They have a wide variety of gel colors to choose from. The prices are very reasonable and they do a great job. The nail techs are very nice and do great work. \vspace{0.5em} \\ \hline
    Copycat & \vspace{0.5em} This is the best nail salon I have ever been to. Everyone is very friendly and professional. My nails look great and I'm glad I did! I will definitely be coming back to this place.\vspace{0.5em} \\ \hline
    MeanSum & \vspace{0.5em} The owner is so nice and accommodating. I went to get my nails done by a friend, and I was extremely happy with this salon. Everyone was very friendly and I was able to use them for nails. They did a great job on my nails and the best part about it was that it was a busy day but it was a treat! Highly recommend them.\vspace{0.5em} \\ \hline
    Lexrank & \vspace{0.5em} I really enjoy coming here to get my nails done. B did an amazing job on my nails. Amazing service and nails. However B did an AMAZING job on my coffin chrome nails and Nancy was extremely helpful figuring out how I wanted my nails done too. Everyone is so friendly there too.\vspace{0.5em} \\ \hline
     \thickhline
    Review 1 & \vspace{0.5em} Tim and Tami always always always have the best customer service and do the best nails. I will NEVER go anywhere else. Even after weeks my nails look and feel as good as they did when I first got them done! I'm so dedicated I recommend and bring in all my friends!\vspace{0.5em} \\ \hline
    Review 2 & \vspace{0.5em} Definitely my new nail salon! Everyone is so friendly and kind, I felt so welcomed! B did an amazing job on my nails. He made sure everything was perfect and happily changed something to make me happy. I would highly recommend this place to anyone who wants A + work at a totally affordable price. Love it!!:)\vspace{0.5em} \\ \hline
    Review 3 & \vspace{0.5em} Amazing service and nails. This is the second time I have been here, they did a perfect job again. They get it done fast yet with precision. Everyone is so friendly there too. Best nail salon I have ever been too. I'm glad I found it.\vspace{0.5em} \\ \hline
    Review 4 & \vspace{0.5em} I really enjoy coming here to get my nails done. They do a wonderful job on both pedis and nails. It is nice and clean inside. They are very friendly and welcoming. It is worth it to stop in and try it out.\vspace{0.5em} \\ \hline
    Review 5 & \vspace{0.5em} My first set of acrylics ever... I decided 27 years was a lot enough time to wait, and I'm SO happy with them. I'm not a huge nail person, and was glad to stumble upon this salon. My nail tech was quiet, clean, and very detail-oriented. Very pleased with my experience here and I recommend this place.\vspace{0.5em} \\ \hline
    Review 6 & \vspace{0.5em} I called to make an appointment for later today for 3 adults and 2 kids and the man who answered the phone said 'we only have 2 techs today' we can't do that. Poor customer service and I never even went in.\vspace{0.5em} \\ \hline
    Review 7 & \vspace{0.5em} Golden Nails has been my nail place for almost a year so it was surprising to see new management. However B did an AMAZING job on my coffin chrome nails and Nancy was extremely helpful figuring out how I wanted my nails done too. Definitely excited to keep coming back!\vspace{0.5em} \\ \hline
    Review 8 & \vspace{0.5em} Seriously the best service I have ever gotten at a Tempe nail salon!! I walked in and they helped me right away. Nancy helped me pick the perfect color and was very honest and up front about everything! I wanted something very natural and using the dip method, I love my nails!! \vspace{0.5em} \\ \thickhline
    \end{tabular}
    \caption{Example summaries produced by different systems on Yelp data.}
\end{table*}


\begin{table*}
    \centering
 	\footnotesize 
    \begin{tabular}{  >{\centering\arraybackslash} m{1.5cm} m{12cm}}
 \thickhline
    Gold & \vspace{0.5em} These are a very comfortable and cute sandal. This thong sandal goes with a variety of outfits and the cushy sole allows for all day comfort. However, they do run a little small, sizing up provides a better fit. Overall, a reasonably priced shoe that will last for years to come.\vspace{0.5em} \\ \hline
    Ours & \vspace{0.5em} These sandals are very cute and comfortable. They fit true to size and are very comfortable to wear. They look great with a variety of outfits and can be dressed up or down depending on the occasion.\vspace{0.5em} \\ \hline
    Copycat & \vspace{0.5em} I love these sandals.They are very comfortable and I can wear them all day without any discomfort. I wear them to work and they are comfortable to wear. \vspace{0.5em} \\ \hline
    MeanSum & \vspace{0.5em} I love these shoes. They are so comfortable and I love the style. They are very comfortable and the perfect price! I would definitely recommend this product to anyone. They are comfortable and stylish.\vspace{0.5em} \\ \hline
    Lexrank & \vspace{0.5em} I have been wearing White Mountain beaded sandals for a couple of years now and they are wonderful. I will never buy from white mountain again. I love White Mountain sandels. Lots of compliments every time I wear them.\vspace{0.5em} \\ \hline
     \thickhline
    Review 1 & \vspace{0.5em} I get constant compliments on these sandals. I order them every summer in a variety of colors. I had heel spurs and back problems so the cushy softness of these is the only thing I can wear comfortably and the small wedge heel is perfect for my back.\vspace{0.5em} \\ \hline
    Review 2 & \vspace{0.5em} These thongs are fun, festive, flexible and surprisingly comfortable. I have very sensitive feet and I can wear these cuties all day. The arch support is great and there is a nice give in the sole. I love these so much I want to put a few pairs away in case they discontinue them. They go with everything.\vspace{0.5em} \\ \hline
    Review 3 & \vspace{0.5em} I have been wearing White Mountain beaded sandals for a couple of years now and they are wonderful. They are lightweight and cushion the feet when worn for long hours. They are also beautiful and usually hold up for two or more seasons.\vspace{0.5em} \\ \hline
    Review 4 & \vspace{0.5em} This was great price for this cute sandal. Unfortunately, the toe piece was very hard and they were a little narrow... unusual since I normally wear a B width. For the right person, they would probably be fine. They just didn't work for me.\vspace{0.5em} \\ \hline
    Review 5 & \vspace{0.5em} I love White Mountain sandels. this is my 2nd pair of these shoes. I wore out the last pair after 2yrs. They are very very blingy and I like that. Would I order another pair? you bet I would / will.\vspace{0.5em} \\ \hline
    Review 6 & \vspace{0.5em} Item was too small, purchased for a friend their size 9 is smaller than the size 8 in the store. Sent it to the wrong address, and I can not seem to find anyone that will tell me where my bill is. I will never buy from white mountain again.\vspace{0.5em} \\ \hline
    Review 7 & \vspace{0.5em} I lived in sandals that looked exactly like this but I thought they were by Bjorn. I couldn't find them anywhere, but found these (go figure). While they aren't quite as comfy as my other ones, I think with a bit of breaking in they'll be just fine. Lots of compliments every time I wear them.\vspace{0.5em} \\ \hline
    Review 8 & \vspace{0.5em} Not only are these SUPER comfortable (yes, even between your toes), they look great with just about anything I wear! I have been complimented on these daily!! I typically wear a 6 1 / 2, I ordered a 7 and they fit perfect!! I need more of these!! HIGHLY RECOMMENDED!!!\vspace{0.5em} \\ \thickhline
    \end{tabular}
    \caption{Example summaries produced by different systems on Amazon data.}
\end{table*}


\begin{table*}
    \centering
 	\footnotesize 
    \begin{tabular}{  >{\centering\arraybackslash} m{1.5cm} m{12cm}}
 \thickhline
    Gold & \vspace{0.5em}This is a perfect compact table that fits well in many places. The chairs are surprisingly very comfortable as well. It is cute and perfect for smaller living quarters and the best part is assembly is simple and straightforward.\vspace{0.5em} \\ \hline
    Ours & \vspace{0.5em} This is a very nice table set for the price. It was easy to assemble and looks great in the kitchen. The only problem is that it is not sturdy enough to hold a lot of weight. It would be nice if it had a little more weight to it so that it would not tip over.\vspace{0.5em} \\ \hline
    Copycat & \vspace{0.5em} This is a great table set for the price. It was easy to put together and looks great. The only thing is that the chairs are a little flimsy, but they are easy to assemble. \vspace{0.5em} \\ \hline
    MeanSum & \vspace{0.5em} The table was very easy to assemble and was easy to assemble. The only thing I would say is that the box is very small and not very sturdy. The table is very sturdy. I would recommend it to anyone looking for a sturdy table and to put on the wall.\vspace{0.5em} \\ \hline
    Lexrank & \vspace{0.5em} The table and chairs are very nice but not quite the color I expected (but I am getting used to it). The table and chairs are solid and sturdy! I received this table and chairs completely damaged. Table and chairs delivered by the carrier right on time and with no damage.\vspace{0.5em} \\ \hline
     \thickhline
    Review 1 & \vspace{0.5em} It was easy to put together and looks great. However, when the item was shipped to me, one of the backs of the chairs was broken. I just fixed it myself with wood glue. Its not even visible now. The rest of it was in perfect condition.\vspace{0.5em} \\ \hline
    Review 2 & \vspace{0.5em} The table and chairs are very nice but not quite the color I expected (but I am getting used to it). Table and chairs delivered by the carrier right on time and with no damage. Very easy to assemble, but very difficult to get out of the box it was so well protected.\vspace{0.5em} \\ \hline
    Review 3 & \vspace{0.5em} This table was super easy to put together. The table and chairs are solid and sturdy! The seats are very comfortable. The table is the perfect size for our not so big kitchen. We are very pleased with this purchase.\vspace{0.5em} \\ \hline
    Review 4 & \vspace{0.5em} Moved to smaller living quarters and this just fits the bill. Color is perfect and it was easy to assemble. One fault to find is that the top scratches easily. It even came with a scratch. Other than that it is fine.\vspace{0.5em} \\ \hline
    Review 5 & \vspace{0.5em} I love my new dining room set. The set is very sturdy, the walnut finish is a nice color.This set is great for a small area, kitchen nook.Would not recommend for a large eating area.Table is small and so are the chairs.Yet strong enough to hold big boys and girls, thumps up, great price, packed well, arrived in a timely matter.\vspace{0.5em} \\ \hline
    Review 6 & \vspace{0.5em} It fits perfectly in the kitchen at the office. My staff assembled it without any delay. Everyone loves the dining set and they can't believe I ordered it on-line. I made the measurements and made sure of the dimensions of the room and the dining set and it's a perfect fit.\vspace{0.5em} \\ \hline
    Review 7 & \vspace{0.5em} I received this table and chairs completely damaged. The customer service experience with this company was terrible. In my opinion, this set is cheap and overpriced. It's not durable and not worth the money. Don't waste your time.\vspace{0.5em} \\ \hline
    Review 8 & \vspace{0.5em} The box looked like it had been opened, and then re-taped for resell. One of the chairs was broken, and the broken piece was nowhere close to the originating piece. Possibly other pieces damaged too, though didn't bother looking, instead just re-taped it back up to be sent back. I hope they don't just resell it to someone else.\vspace{0.5em} \\ \thickhline
    \end{tabular}
    \caption{Example summaries produced by different systems on Amazon data.}
\end{table*}

\end{document}